\documentclass[letterpaper, 10 pt, journal, twoside]{ieeetran} 
\ifCLASSINFOpdf
\else
\fi

\usepackage[english]{babel}
\usepackage{amsmath}
\usepackage{amssymb}
\usepackage{multirow}
\usepackage{rotating}
\usepackage{multirow}
\usepackage{mathtools}
\usepackage{units}
\usepackage{subfloat}
\usepackage{float}
\usepackage[]{units}
\usepackage{xcolor}
\usepackage{todonotes}
\usepackage{hyperref}

\usepackage{algpseudocode}
\usepackage{algorithm}
\usepackage{graphicx}
\usepackage{caption}
\usepackage{subfigure}
\usepackage{verbatim} 

\newcommand{\vect}[1]{\boldsymbol{#1}}

\definecolor{CommentDG}{rgb}{0,0.6,0}
\newcommand{\iks}[1] {{\color{CommentDG} { IKS: \textbf{#1}}}}

\begin{document}
%
\title{Control of a Quadrotor with Reinforcement Learning}
%
%
%

\author{Jemin Hwangbo$^{1}$, Inkyu Sa$^{2}$, Roland Siegwart$^{2}$ and Marco Hutter$^{1}$
\thanks{Manuscript received: February, 16th, 2017; Revised May, 11th, 2017; Accepted June, 8th, 2017.}
\thanks{This paper was recommended for publication by Editor Tamim Asfour upon evaluation of the Associate Editor and Reviewers' comments. *This work was funded by Swiss National Science Foundation (SNF) through the National Centre of Competence in Research Robotics and Project 200021-166232. This project also has received funding from the European Union’s Horizon 2020 research and innovation programme under grant agreement No 644227 and from the Swiss State Secretariat for Education, Research and Innovation (SERI) under contract number 15.0029.}
\thanks{$^{1}$RSL, ETH Zurich, Switzerland
        {\tt\small jhwangbo@ethz.ch}}%
\thanks{$^{2}$ASL, ETH Zurich, Switzerland }%
\thanks{Digital Object Identifier (DOI): see top of this page.}
}


%
%

\markboth{IEEE Robotics and Automation Letters. Preprint Version. June, 2017}
{Hwangbo \MakeLowercase{\textit{et al.}}: Quadrotor control with RL}  
%



\maketitle

\begin{abstract}
In this paper, we present a method to control a quadrotor with a neural network trained using reinforcement learning techniques. With reinforcement learning, a common network can be trained to directly map state to actuator command making any predefined control structure obsolete for training. Moreover, we present a new learning algorithm which differs from the existing ones in certain aspects. Our algorithm is conservative but stable for complicated tasks. We found that it is more applicable to controlling a quadrotor than existing algorithms. We demonstrate the performance of the trained policy both in simulation and with a real quadrotor. Experiments show that our policy network can react to step response relatively accurately. With the same policy, we also demonstrate that we can stabilize the quadrotor in the air even under very harsh initialization (manually throwing it upside-down in the air with an initial velocity of \unit[5]{m/s}). Computation time of evaluating the policy is only \unit[7]{$\mu$s} per time step which is two orders of magnitude less than common trajectory optimization algorithms with an approximated model.
\end{abstract}

\section{Introduction}

\IEEEPARstart{R}{einforcement} learning is successful in solving many complicated problems. Its advantage over optimization approaches and guided policy search methods~\cite{hwangbo2015direct} is that it does not need a predefined controller structure which limits the performance of the agent and costs more human effort. Recent works (e.g. \cite{alphago,Atari}) show that well trained networks perform even better than human experts in many complicated tasks. They also show promising results in learning tasks with continuous state/action space~\cite{DDPG} which are closely related to robotics. Although reinforcement learning has been used in robotics for many decades, it has largely been confined to higher-level decisions~(e.g. trajectory) rather than low actuator commands. In this work, we demonstrate that an aerial vehicle can be fully controlled using a neural network which was trained in simulation using reinforcement learning techniques. Our policy network is a function directly mapping a state to rotor thrusts so there are only a few assumptions made with respect to the structure of the controller. This proves that a unifying control structure for many robotics tasks is possible.

Policy learning on an aerial vehicle is often demonstrated in literature. Guided policy search with a MPC controller~\cite{immitationLearningMPC} is demonstrated in simulation. This work uses a policy that maps the raw sensor data to the rotor velocities. Quadrotor control with reinforcement learning policy is demonstrated in~\cite{RLquadrotor} with a real flying vehicle. The authors used model-based reinforcement learning to train a locally-weighted linear regression policy. They achieved a limited amount of success in controlling a quadrotor for a step response and hovering motion. In this work, we show more dynamic motion (i.e. dynamic stabilization from an upside-down throws) can be achieved with reinforcement learning.

We also introduce a new learning algorithm that we used to train a quadrotor. The new algorithm is a deterministic on-policy method which is not common in reinforcement learning. We demonstrate that, using zero-bias, zero-variance samples, we can stably learn a high-performance policy for a quadrotor. In addition, due to the fact that we are using small number of high quality samples, there is only a small burden in neural network computation compared to many state-of-the-art algorithms. This makes our method very practical for optimization in simulation where network-related computations are usually heavier than dynamic simulation.
\begin{figure}[t]
\centering 
\includegraphics[width=0.48\textwidth]{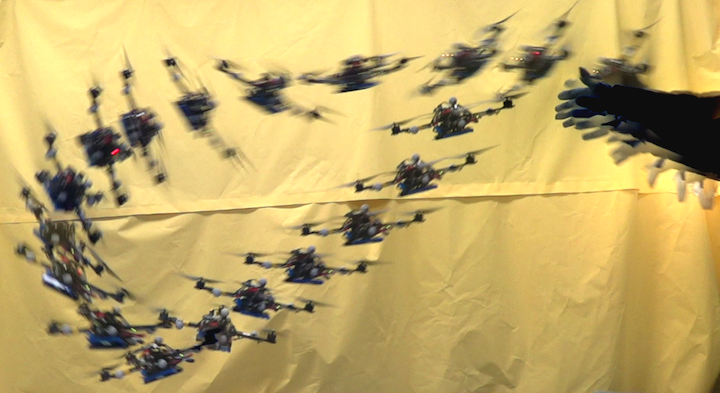}
\caption{The quadrotor stabilizes from a hand throw. The motor was enabled after it left the hand}\label{fig:frontImg}
\vspace{-0.4cm}
\end{figure}
We also present in detail what dynamic model is used, how we set up the problem and how the learning is performed. We demonstrate the performance of the trained policy both in simulation and on the real flying vehicle. In simulation, we demonstrate recovery from random initial states (i.e. random twist and pose). The same controller is tested on the real hardware for waypoint tracking and recovery from manual throws. We present the results from all tests and compare the differences between the results from simulation and those from the real world.

The advantages of neural network policies are not limited to their versatility. It is extremely cheap to evaluate due to their approximated representation. The computation time to evaluate the policy in the present example is only \unit[7]{$\mu$s}~(measured with a single core of Intel Xeon E5-1620). This offers more computational resources for other algorithms running on the robot. It can be easily extended to combine other functionalities~(e.g. state estimation and object detection) but they lie outside of the scope of this project.

The remainder of this paper is structured as follows. Section \ref{sec:bg} introduces related works. Section \ref{sec:Method} describes the proposed value and policy networks, their exploration strategy, and training approach. We demonstrate our simulation and experimental results in Section \ref{sec:exp} and conclusions are drawn in Section \ref{sec:conclusion}.

\section{Background}\label{sec:bg}

The presented approach is built on deterministic policy optimization using a natural gradient descent~\cite{ngd}. Deterministic policy optimization has three main advantages over stochastic policy optimization. Firstly, value/advantage estimate from on-policy samples have lower variance (zero when the system dynamics is deterministic), which makes the learning process more stable. Secondly, it is possible to write the policy gradient in a simpler form which makes it computationally more attractive. Lastly, we do not want the quadrotor to be controlled by a stochastic policy, since this can lead to poor and unpredictable performance. 

On the contrary, a deterministic policy gradient method requires a good exploration strategy since, unlike stochastic policy gradient, it has no clear rule for exploring the state space. In addition, stochastic policy gradient methods tend to solve more broad classes of problems from our experience. We suspect that this is due to the fact that stochastic policy gradient has less local optima that are present in deterministic policy gradient.

In reinforcement learning, we obtain samples from a black-box system or from a real robot of which we do not assume any model. Starting from an initial state which is distributed according to $d_0(s)$, we choose a series of actions $a \in \mathcal{A}$ for \unit[$T$]{steps} in order to obtain a trajectory $(s_{1:T+1}, a_{1:T}, r_{1:T})$ where $s \in \mathcal{S}$ is the state and $r\in \mathbb{R}$ is the value of the deterministic cost function $R: \mathcal{S} \times \mathcal{A} \rightarrow \mathbb{R}$. Our goal is to find a parameterized policy $\pi_\theta$, where $\theta$ is called policy parameter, which minimizes the average value over states
\begin{equation}
\begin{split}
L(\pi_\theta) &= \mathop{\mathbb{E}}_{s_0, s_{1}...}[\sum_{k=0}^{\infty}\gamma^k r_{t+k}] = \int_{\mathcal{S}}d^{\pi_\theta}(s)V^{\pi_\theta}(s)ds,\\
\text{where, }&~V^{\pi_\theta}(s) = \mathop{\mathbb{E}}_{s_{t+1}, s_{t+2}...}[\sum_{t=t}^{\infty}\gamma^t r_t|s_t=s, \pi_\theta].
\end{split}
\end{equation}
It describes the averaged value over the stable state distribution of policy $\pi$ assuming that such distribution exists. We also assume that the discount factor $\gamma \in [0,1)$ limits the values of the value function $V$ to be finite such that it is well-defined.

According to~\cite{dpg}, a deterministic policy gradient w.r.t. the policy parameters exists. For simplicity, we ignore the discount in the state distribution and write the gradient as
\begin{equation}
\nabla_\theta L(\pi_\theta) = \mathop{\mathbb{E}}_{s\sim d^{\pi_\theta}(s)}[\nabla_\theta \pi_\theta(s) \nabla_a Q^{\pi_\theta}(s,a)| a=\pi_\theta(s)],
\end{equation}
where $Q^{\pi_\theta}(s_t,a_t) = \mathop{\mathbb{E}}_{s_{t+1}}[r(s_t,a_t)+\gamma V^{\pi_\theta}(s_{t+1})]$ is called action-value function. Its output can be interpreted as a value of taking a particular action at a particular state and following the policy thereafter. We use the baseline function $V^{\pi_\theta}(s)$ and rewrite the policy gradient as
\begin{equation}
\nabla_\theta L(\pi_\theta) = \mathop{\mathbb{E}}_{s\sim d^{\pi_\theta}(s)}[\nabla_\theta \pi_\theta(s) \nabla_a A^{\pi_\theta}(s,a)| a=\pi_\theta(s)],
\end{equation}
where $A^{\pi_\theta}(s,a)=Q^{\pi_\theta}(s,a)-V^{\pi_\theta}(s)$ is called advantage function, whose value can be interpreted an advantage in value gained by taking a certain action over the action from the current policy $\pi_\theta$(s).

\section{Method}\label{sec:Method}
This section describes the method that we used to train our policy for a quadrotor. The validity of this method on other tasks should be further analyzed in the future.

\subsection{Network Structure}~\label{sec:netStruc}
There are two networks used for training, namely a value network and a policy network. Both networks have the state as an input. We used nine elements of the rotation matrix $R_b$ to represent the rotation and the rest of the states are trivially represented by position, linear velocity and angular velocity of the system, with adequate scaling that makes the states roughly follow a normal distribution. A more common rotation parameterization method is a unit quaternion which has a certain pitfall in our case. It is that there are two values representing the same rotation~(i.e. $q=-q$), thus either requiring double the training data or end up with a discontinuous function when we limit our domain to one of the hemispheres of $S^3$. The rotation matrix is a highly redundant representation but simple and free from such pitfall.

\begin{figure}
\begin{center}
\subfigure[Policy network.]{\includegraphics[width=\columnwidth]{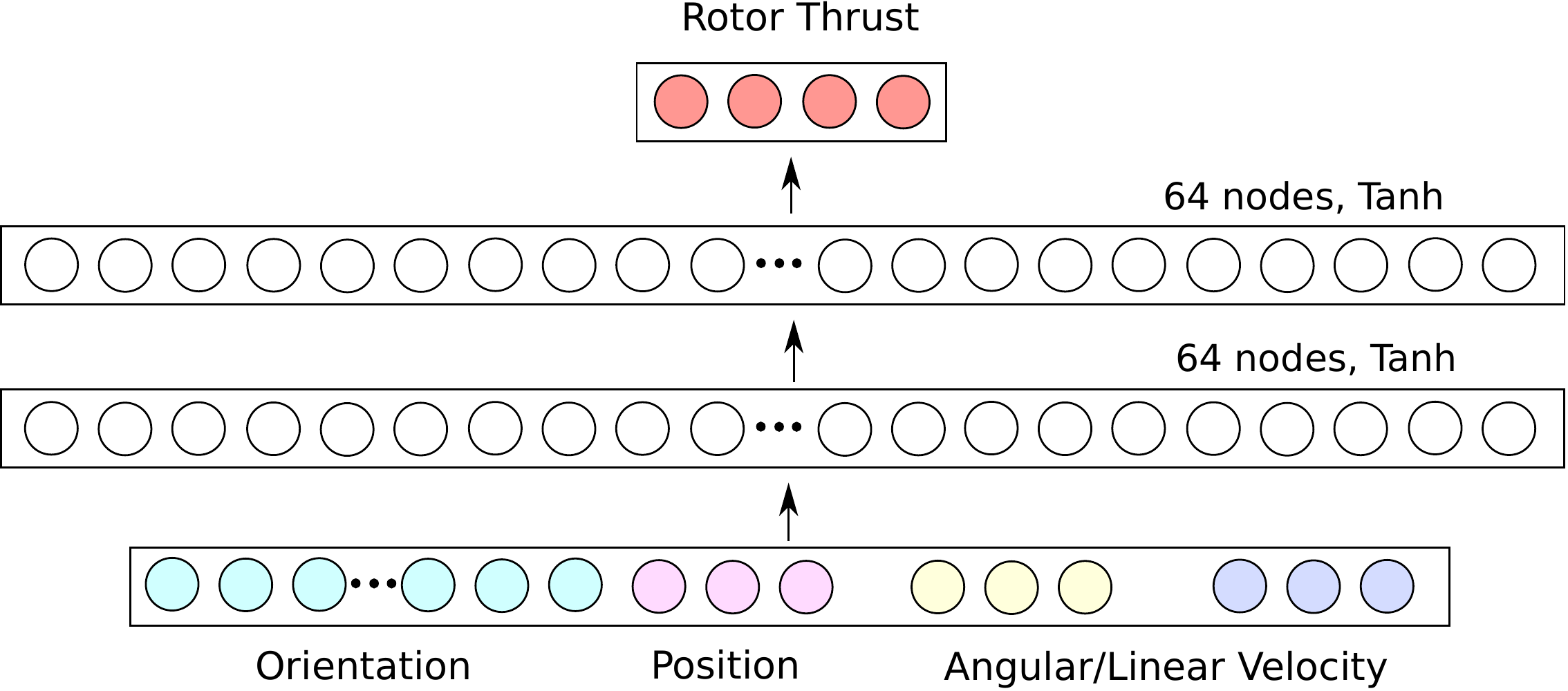}}
\subfigure[Value network.]{\includegraphics[width=\columnwidth]{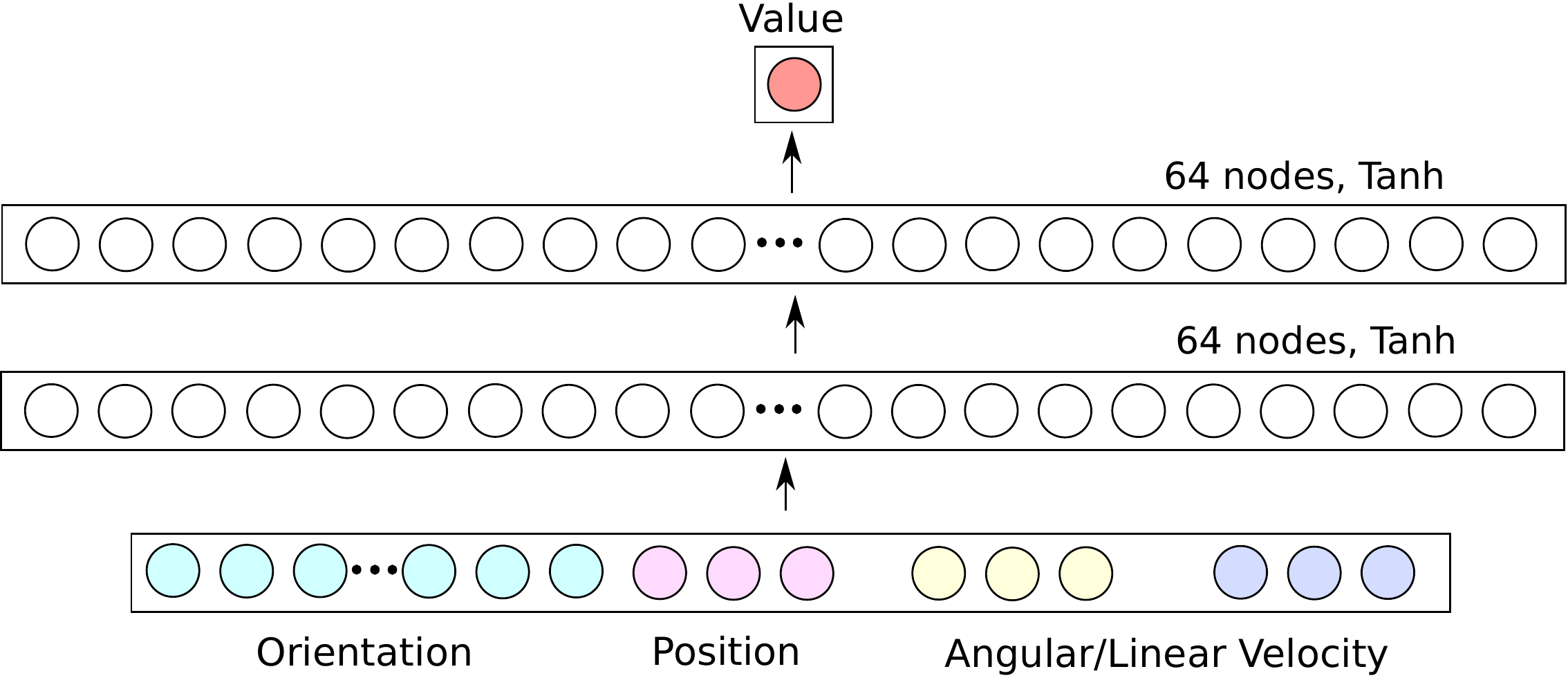}}
\end{center}
\vspace{-15pt}
	\caption{The two neural networks used in this work are shown.}
	\label{fig:networks}
\vspace{-10pt}
\end{figure}

Consequently, we have a 18-dimensional state vector and a 4-dimensional action vector. We use 2 hidden layers of 64 \texttt{tanh} nodes for each. The structures are illustrated in Fig~\ref{fig:networks}. The structure is not optimized in any sense. In fact, we did not try different number of nodes and layers. From our experience, neural networks are quite versatile and can cope with variety of problems with a single structure.

\subsection{Exploration Strategy}\label{sec:expStra}
\begin{figure}[t]
\vspace{0.2cm}
\centering 
\includegraphics[width=0.48\textwidth]{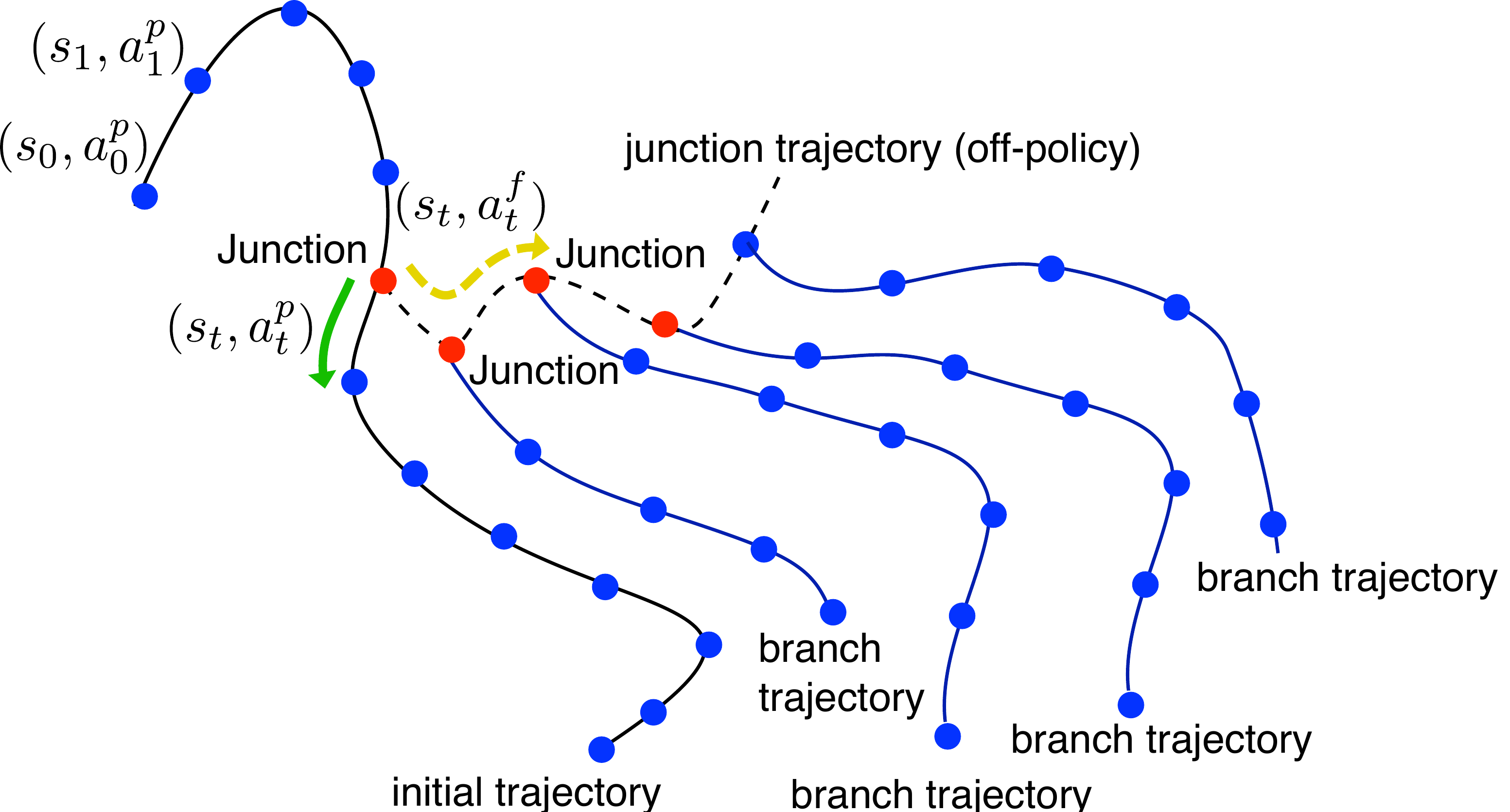}
\caption{Exploration strategy.}\label{fig:exp}
\end{figure}
We consider a simple exploration strategy that is described in~\cite{vine,TRPO} as shown in Figure~\ref{fig:exp}. The trajectories are separated into three categories: initial trajectories, junction trajectories and branch trajectories. The initial and branch trajectories are on-policy and the junction trajectories are off-policy generated with an additive Gaussian noise with covariance $\Sigma$. The branch trajectories are on-policy trajectories starting from some state along the junction trajectories. The idea here is to get an unbiased advantage/value estimation when both the policy and the environment are deterministic. The motivation of having junction trajectories longer than one time step is to get more broadly distributed samples, since the borders of the sampling region are usually not well approximated with neural networks. Too long junction trajectories mean that our assumption that junctions are distributed according to $d^\pi(s)$ is violated. However, it does not affect the performance in practice if the junction trajectories are still far shorter than initial trajectories. In addition, using the new broad distribution make it less prone to being trapped in a local minimum in some problems. In our problem, since the random initialization solved the state exploration problem, both one-step and multi-step junction trajectories were performing similarly. The length of the junction trajectories is a tuning parameter for different problems when a single step junction trajectories are not sufficient.

Note that there are many simulations running in one core of CPU but they are synchronized such that we can forward evaluate the policy network with a batch of states. We chose this setup since the actual computation time of the simulation was relatively short and minimizing the network calls significantly reduced the computation time. This approach can optimize the use of the GPU\footnote{one NVIDIA GeForce Titan X (Maxwell), used for training.}.

In practice, since all trajectories have a finite length, the tail costs (i.e. value of the terminal states) are estimated using the approximated value function $V(s|\eta)$, where $\eta$ is the value function parameter vector. Longer branch trajectories means that the learning step requires more evaluations per iteration but the estimate has lower bias. The quadrotor simulation is noiseless and there is zero variance with our deterministic policy. So advantage estimates from longer trajectories are always more accurate. Since our focus is not on fast convergence but rather on stable and reliable convergence, we use long trajectories in this work. For noisy systems, adequate lengths for the trajectories have to be found. Alternatively, we can draw on a general advantage estimation method~\cite{TRPOgae} to improve the performance.

\subsection{Value Function Training}
The value function is trained using Monte-Carlo samples that are obtained from on-policy trajectories. Since the trajectories have finite length, we obtain the terminal value from the current value function. Mathematically, we can write it as
\begin{equation}\label{equ:value}
v_i = \sum_{t=i}^{T-1}\gamma^{t-i} r^{p}_t + \gamma^{T-i}V(s_{T}|\eta),
\end{equation}
where $\eta$ is the parameters of the approximated value function and $T$ is the length of the trajectory. When the system has no noise, assuming we only update at the end of episodes, this method is always better than temporal difference learning or TD($\lambda$)~\cite{SuttonBook}.  We are exploiting the fact that our system is deterministic. TD($\lambda$) can be superior with noisy systems when $\lambda$ is well tuned.

We use all the states from on-policy trajectories. We run the optimizer for 200 iterations per learning step but terminate if the loss goes below 0.0001. Instead of squared error function, we use Huber loss~\cite{huberLoss}.

\subsection{Policy Optimization}
We perform policy optimization using natural gradient descent. The common way to define a distance measure in stochastic policy optimization is with average Kullback–Leibler~(KL) divergence~\cite{klDivergence}, which describes the distance between two distributions. The respective Hessian~(the first order is zero) is given by the Fisher Information Matrix~(FIM) which is a common metric tensor in the policy parameter space. Since we want to describe the distance between the sample distribution and the new deterministic policy, an intuitive alternative is the Mahalanobis metric. In our setup, we use an analytical measure, which describes the distance between the action distribution and our new policy, instead of using the sample distribution for computational simplicity. We define our policy optimization as following:
\begin{equation}\label{equ:policyOpt}
\begin{split}
&A^\pi(s_i, a^f_i) = r_i^f + \gamma v_{i+1}^f - v_i^p,\\
&\bar{L}(\theta) = \sum^K_{k=0} A^\pi(s_k,\pi(s_k|\theta)),\\
&\theta_{j+1} \leftarrow \theta_j - \frac{\alpha}{K} \sum^K_{k=0} n_k,\\
&s.t.~(\alpha n_k)^TD_{\theta \theta}(\alpha n_k) < \delta,~~\forall k,
\end{split}
\end{equation}
where $n_k$ is per-sample natural gradient defined as a vector that satisfies $D_{\theta\theta}n_k=g_k$, where $D$ is the squared Mahalanobis distance and the double subscript means that it is a Hessian w.r.t. the subscripted variable. We use $i$ for time index, $k$ for junction index, and $j$ for iteration index. We denote the on-policy transitions with superscript $p$ and the off-policy transitions with a superscript $f$ to avoid confusion with branching trajectories. The bar denotes that $\bar{L}$ is an approximation of $L$ from samples which are sampled from the distribution $d^\pi(s)$. 

It is not trivial to find $d\bar{L}(\theta)/da$ since we only have a discrete samples of $A^\pi$ rather than the model. We use a linear model of the two points as an approximation, which yields $g_k \approx A^\pi_k(a^f_k-a^p_k)/||(a^f_k-a^p_k)||^2$. The inequality constraint is called trust region constraint which limits the update contribution of each sample, since our gradient estimate might be extremely large for a small noise vector $(a^f_k-a^p_k)$.

The inverse of $D_{aa}$ maps the policy gradient w.r.t. action to the natural gradient w.r.t. action. Such mapping is one-to-one~(i.e. both $D_{aa}$ and $D_{aa}^{-1}$ exist) as long as the covariance is full rank. However since we are interested in the Hessian w.r.t. the policy parameters, $D_{\theta\theta}$ is not full rank and we cannot find its inverse. TRPO~\cite{TRPO} use conjugate gradient to overcome the problem of finding the inverse Hessian matrix explicitly, but this method gives only an approximate solution. We use the Singular Value Decomposition~(SVD) method to find the pseudoinverse of the Hessian matrix which gives an exact solution. Note that since the per-sample gradient $g_k = \frac{\partial A}{\partial a}\frac{\partial a}{\partial \theta}$ lives in the support of the Hessian matrix $H_{\theta \theta} = \frac{\partial a}{\partial \theta}^TD_{aa}\frac{\partial a}{\partial \theta} = J^T D_{aa} J$, the linear equation $H_{\theta\theta}n_k=g_k$ has a solution and it can be obtained by pseudoinverse\footnote{given $rank(J) \geq |a|$, where $|\cdot|$ is the cardinality function. It is almost always satisfied with a neural network and we assume that it is true.}.

Since directly computing the pseudoinverse of the Hessian $H_{\theta \theta}$ is prohibitively expensive for neural networks, we use the following algebraic tricks:
\begin{equation}
\begin{split}
H_{\theta \theta}^+ &= (J^T D_{aa}J)^+ = (J^TL_{aa}L_{aa}^TJ)^+\\
& = (L_{aa}^TJ)^+((L_{aa}^TJ)^+)^T = V\Sigma_v^+ U^TU\Sigma_v^+ V^T\\
&= V(\Sigma_v^+)^2 V^T,
\end{split}
\end{equation}
where $L_{aa}$ is a Cholesky factor of $D_{aa}$. Since $D_{aa} = \Sigma^{-1}$, it is symmetric and positive-definite and hence the Cholesky factor exists. We use thin SVD, $L_{aa}^TJ =V\Sigma_v U^T$, to simplify the computation. Thin SVD only finds the non-zero singular values and their corresponding blocks of $U$ and $V$ such that $\Sigma_{v}$ is a square matrix and $V$ has the same dimension as $J$. Note the notational difference between the singular value matrix $\Sigma_v$ and  our noise covariance $\Sigma$.  The computation time of $\Sigma_v^+$ is negligible because the operation is just an element-wise inverse. Hence only SVD is computationally costly in this formulation. Even SVD has favorable computational complexity $O(min(mn^2, m^2n))$, i.e. square to the cardinality of the action space and linear to the number of parameters, so it is applicable to larger neural networks as well. For the given network structure in sec~\ref{sec:netStruc}, the Cholesky decomposition and the SVD takes about \unit[20]{\%} of the whole policy optimization. Our benchmark test shows that it takes about \unit[0.35]{ms} per sample while conjugate gradient with 10 iterations takes \unit[1.1]{ms} for the given Jacobian used in this work which is a 4$\times$5636 real matrix. However, in terms of accuracy, conjugate gradient method is also near the exact solution in practice. We did not observe an error more than $10^{-10}$ with conjugate gradient. We believe that the difference is becomes significant when the covariance matrix is ill-conditioned.

In many other algorithms, the noise covariance $\Sigma$ is often updated using a policy gradient. However, $\Sigma$ is for exploration and not a variable to be optimized in a deterministic policy. An adequate $\Sigma$ is the one that is big enough to allow exploration and whose inverse is proportional to the average metric in the action space. The metric of the action space for policy optimization is intuitively defined as $Q^\pi_{aa}(s, a)$ which, roughly speaking, gives us a sense of the scale of the action. As it is noted, such metric is state-dependent and it is not just a single matrix. However, since it is not practical to have different noise depending on the state, we use a single matrix for noise. We define such noise manually since we usually have a good idea of the scale of the actions. Automatic covariance adjustment method will be an interesting future work in this regard.

The full algorithm is summarized in Alg.~\ref{alg1}

\begin{algorithm}
\caption{Policy optimization}
\begin{algorithmic}[1]
\State Give initial parameters of $V(s|\eta)$ and $\pi(s|\theta)$
\While{$j=1,2,3$ ... until convergence}
\State Collect data according to section~\ref{sec:expStra}
\State Compute MC estimate of $v^p_i$ using Equ.~\ref{equ:value}
\State Update $V(s|\eta)$ for $n_{vs}$ times using Huber loss
\State Update $P(s|\theta)$ once using junction pairs and natural gradient descent as in Equ.~\ref{equ:policyOpt}
\EndWhile
\end{algorithmic}\label{alg1}
\end{algorithm}

\section{Policy Optimization in Simulation}
We developed a software framework called Robotic Artificial Intelligence~(RAI)\footnote{\url{https://bitbucket.org/leggedrobotics/rai}}. In contrast to the existing frameworks (e.g.~\cite{rllab}), RAI is written in C++ for fast computation. One of the big advantages of RAI is that it offers numerous utilities for logging, timing, plotting, 3D animation for visualization and video recording of the simulation. They lead to faster debugging and better analysis on computational resource consumption.

We also used our own code to simulate the quadrotor in order to ensure that it is numerically accurate and stable. Since the simulation is also written in C++, the computation time of integrating dynamics was far less than network training time.

\subsection{Robot model for simulation}
We used a very simple model for the simulation. Note that we do not intend to model every detail of the quadrotor. We ignore all drag forces acting on the body and use a simple floating body model with four thrust forces acting on the body. The equation of the motion can be written as,

\begin{equation}
\vect{J}^T\vect{T} = \vect{M}\vect{a} + \vect{h}
\end{equation}

where $\vect{J}$ is the stacked Jacobian matrices of the centers of the rotors, $\vect{T}$ is the thrust forces, $\vect{M}$ is the inertia matrix, $\vect{a}$ is the generalized acceleration and $\vect{h}$ is the coriolis and gravity effect. The propellers can only produce positive force (upward force on the quadrotor) and we simply threshold the thrust to zero in the simulation whenever we detect negative thrust. Since we only have a single floating body, the equation collapses to Newton-Euler equation. The inertia matrix is block-diagonal matrix for a floating body and the forward dynamic computation is extremely efficient. In fact, we even simplified it to a diagonal matrix since we did not measure the inertia. We use a very simple model to point out that it is often possible to have a good performance even without taking much effort to model many details of physics.

We use a boxplus operator~\cite{boxplus} to improve the accuracy of the integration since the motion of the quadrotor is very dynamic and the simulation might become inaccurate. This let us use a big time step in integrating the dynamics~(\unit[0.01]{s}).

\subsection{Problem Formulation}
We are interested in waypoint tracking with a quadrotor without generating a trajectory. In addition, we want to stabilize the system in any configuration whenever it is physically possible (i.e. upside down with a random linear and angular velocity). During policy optimization, we train it to go to the origin of the inertial frame. During operation, we input the state subtracted by the waypoint location to the policy. This way we do not have to train waypoint tracking explicitly. The quadrotor is initialized in a random state (position, orientation, angular/linear velocities are all random) with a reasonable bound such that we can easily explore the feasible state space.

We added a simple Proportional and Derivative~(PD) controller for attitude with low gains along with our learning policy. The sum of the outputs of the two controllers are used as a command. While training, we noticed that the simulation sometimes become unstable (get NaN in simulation) when the angular velocity becomes very high. We suspect that this is due to the fact that the algorithm initially explores a large region where we cannot simulate accurately. In addition, the gradient observed when the quadrotor is upside-down is very low and discontinuous. This means that the gradient-based algorithms take very long time to learn. 
Note that, as it will be shown in the following section, this PD controller alone is simply insufficient and generates very high costs. The quadrotor simply flies away since it is initialized with a high initial velocity. The PD controller is used to stabilize the learning process but it does not aid the final controller since the final controller manifests much more sophisticated behaviors, as will be shown in the following sections. 

We use a PD controller in the following form:

\begin{equation}
\vect{\tau}_b = \vect{k}_p \vect{R}^T\vect{q} + \vect{k}_d \vect{R}^T\vect{\omega},
\end{equation}

where $\vect{\tau}_b$ is the virtual torque produced on the main body as a result of the thrust forces, $\vect{q}$ and $\vect{R}$ are the orientation of the quadrotor in Euler vector and rotation matrix forms respectively, and $\vect{\omega}$ is the angular velocity. All elements of the controller gains $\vect{k}_p$ and $\vect{k}_d$ are set to $-0.2$ and $-0.06$ respectively except for the z-direction gains which are set to one sixth of the those of other axes. Note that a PD controller on Euler angles is insufficient for us since we explore all orientations including the ones near the singularity. This PD controller ensures that the rotors apply torque in the direction of the minimum path. In addition to the PD controller, we also use a bias on rotor thrust that is just enough to compensate for gravity when the quadrotor is flying in nominal orientation.

The cost is defined as

\begin{equation}
r_t = 4\times10^{\mbox{\footnotesize -3}}||\vect{p}_t|| + 2\times10^{\mbox{\footnotesize -4}}||\vect{a}_t|| + 3\times10^{\mbox{\footnotesize -4}}||\vect{\omega}_t|| + 5\times10^{\mbox{\footnotesize -4}}||\vect{v}_t||,
\end{equation}

where $p_t$, $\omega_t$ and  $v_t$ are position, angular velocities and linear velocities respectively. Only the position has a high cost coefficient since that is what we care about the most. We roughly set the rest of the coefficients such that the other cost terms have about one tenth of the magnitude of the position error. We used a discount factor of $\gamma = 0.99$.

\subsection{Network Training}
\begin{figure}[t]
\centering 
\includegraphics[width=0.47\textwidth]{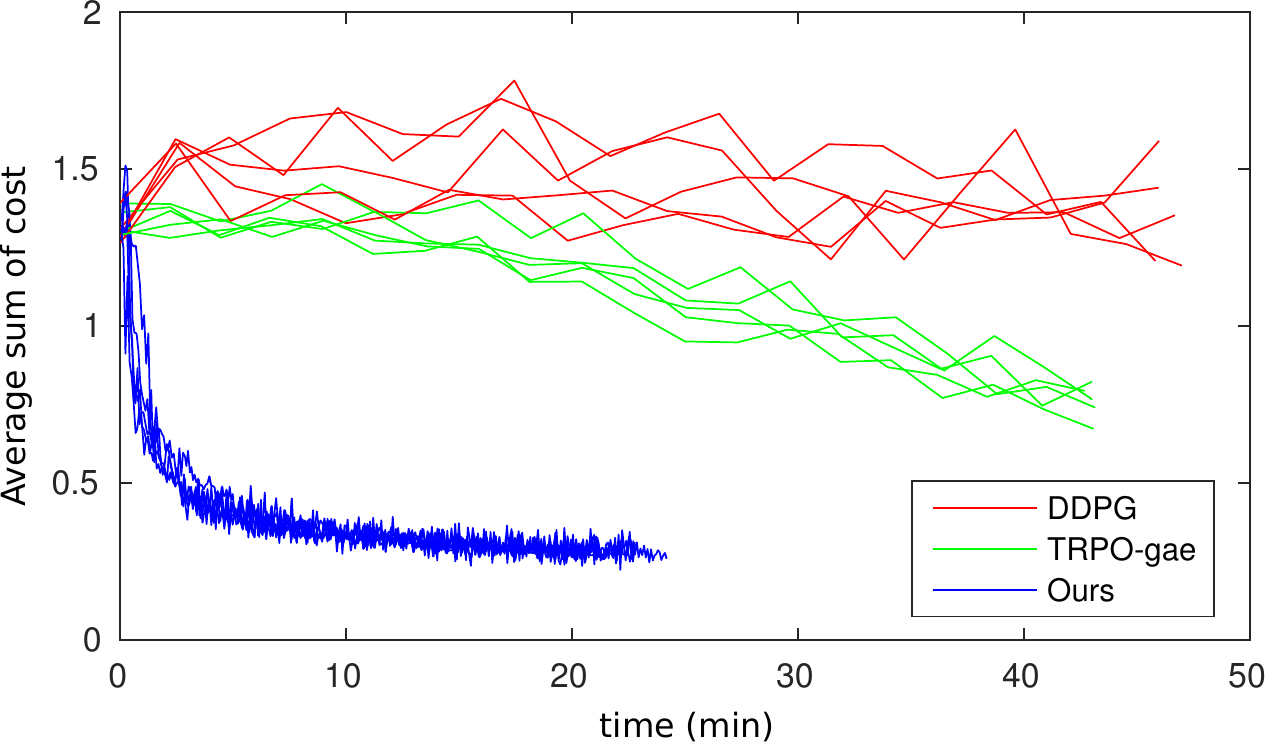}
\caption{Learning curves for optimizing the policy for three different algorithms and 5 different runs per each algorithm.}\label{fig:learningcurve}
\end{figure}

\begin{figure}
\centering 
\includegraphics[width=0.35\textwidth]{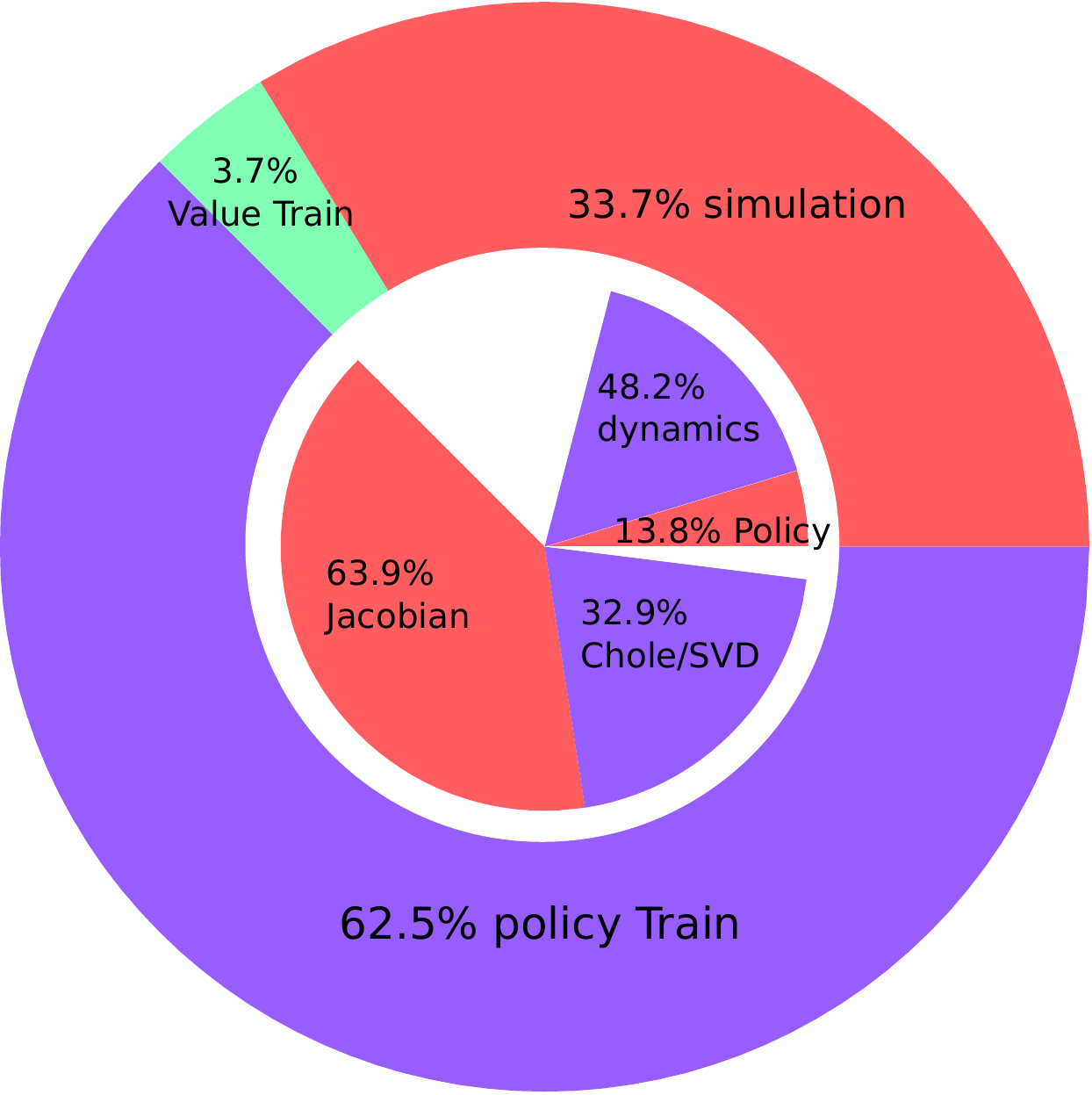}
\caption{Computation resource consumption is shown. The outer ring represents the fraction of each categories and the inner ring is for that of their subcategories.}\label{fig:piechart}
\end{figure}

As described in section~\ref{sec:Method}, we train the value network using the on-policy samples. Since we are not focusing on fast, but rather on stable and reliable convergence, we set the algorithmic parameters to conservative values. We used 512 initial trajectories and 1024 branching trajectories with noise depth of 2 which corresponds to 1.0 million time steps per iteration. Although this number seems high, it took less than ten seconds per iteration due to parallelization of rollouts. Note that our conservative method of getting advantage estimate is sample-expensive but the whole optimization is relatively cheap because it only uses a subset of transition tuples for policy update. We simulate the rollouts in a batch for one time step, collect the states and forward the network in a batch. This was extremely helpful in reducing the learning time. After 10~minutes of training, the performance of the policy was visually good but the average cost value decreased slightly but continuously for 25~minutes. The average performance for the policy was evaluated using 10 rollouts at the end of every iteration and the result is shown in Fig.~\ref{fig:learningcurve}. We also ran the learning task on TRPO and DDPG. DDPG was not able to converge to adequate performance in reasonable amount of time. TRPO managed to reach the same performance (cost of 0.2$\sim$0.25) as the proposed method but for much longer period of time. TRPO and the proposed method performed similarly when compared in performance per simulation time. However, since the simulation time is relatively short compared to the neural network back propagation and the conjugate gradient used in TRPO, the proposed method was far more practical. The computational resource consumption of the proposed method is illustrated in Fig.~\ref{fig:piechart}.

\subsection{Performance in Simulation}

We assessed the stability of the policy by randomly placing the quadrotor in different states and recorded the failure rate. Here the failure means that the quadrotor was touching the ground. We used a linear MPC controller~\cite{2016arXiv161109240K} as a baseline controller for the performance evaluation. The orientation was sampled uniformly in $SO(3)$ and all other quantities are sampled uniformly from $[-1,1]$. The learned policy and MPC policy had a failure rate of \unit[4]{\%} and \unit[71]{\%} respectively for 100 rollouts. As expected, it was impossible to recover from certain initial conditions, e.g. upside-down with full downward speed. However, the overall performance shows that the learned policy is reliable in recovery. Some of the trajectories from simulation with the learned policy will be shown in the later section together with the experimental results.


The policy network code was ported to a series of Matrix arithmetics using the Eigen library and it took about \unit[7]{$\mu$s} to evaluate the policy for a given state. This makes it nearly computation-free. In comparison, the linear MPC controller \cite{kamelmpc2016} takes about \unit[1,000]{$\mu$s} for one time step.

\section{Experiments}\label{sec:exp}
\begin{figure}[H]
\centering
\includegraphics[width=0.47\textwidth]{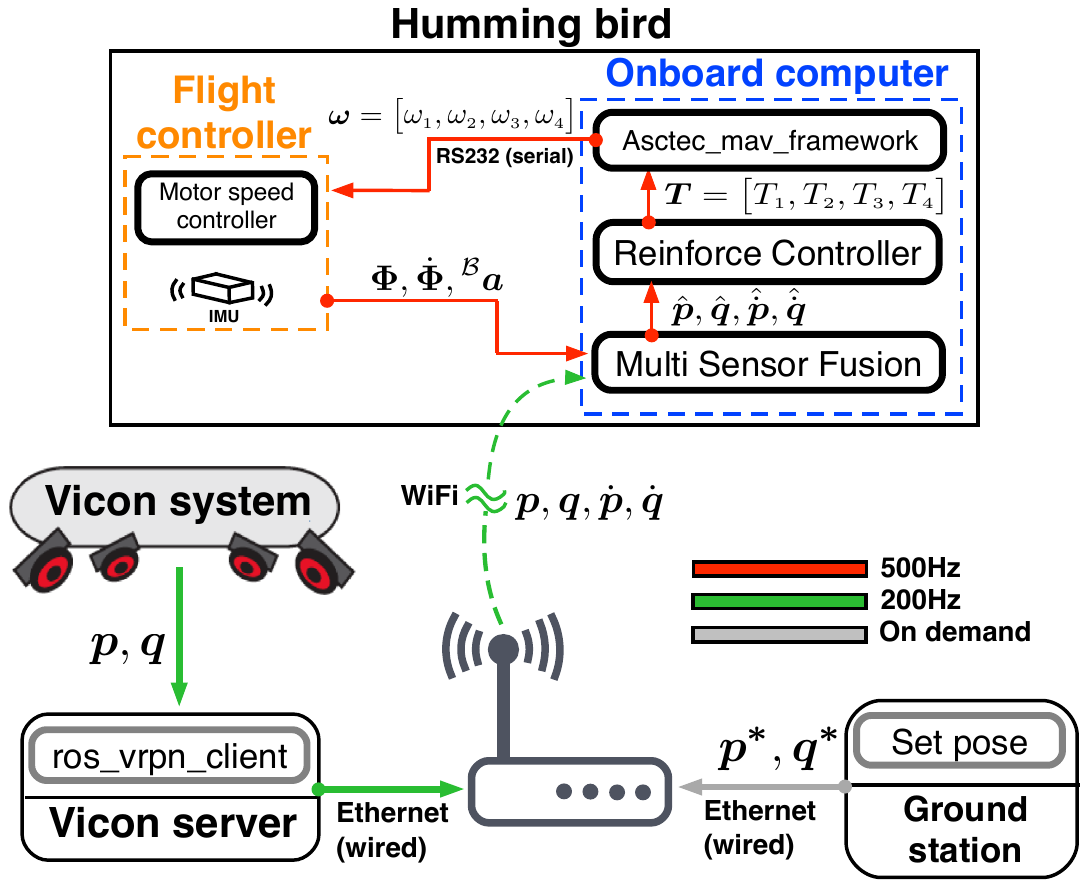}
\caption{System diagram used for the experiments. Different colors denote the corresponding rate respectively. $\boldsymbol{p},\boldsymbol{q},\dot{\boldsymbol{p}}, \dot{\boldsymbol{q}}$ are position, orientation and their velocities. $\boldsymbol{p^{*}}$ and $\hat{\boldsymbol{p}}$ denote the desired goal and estimated position. $\boldsymbol{T}$ and $\boldsymbol{\omega}$ are thrust in \unit{N} and rotor speed in \unit{rad/s}. $\boldsymbol{\Phi},\dot{\boldsymbol{\Phi}},{}^{\mathcal{B}}\boldsymbol{a}$ represent IMU measurements; orientation, angular velocity, and linear acceleration in body frame.}\label{fig:diagram}
\end{figure}

Without further optimization, we evaluated the policy learned in the simulation on the real quadrotor. We used a Hummingbird quadrotor from Ascending Technologies whose physical parameters are listed in Tab.~\ref{tab:pp}. The vehicle carries an Intel Computer Stick that is 0.059\unit{kg} and has 1.44\unit{GHz} quad-core Atom processors for onboard calculation. Note that we used the same model parameters in the simulation. A Vicon motion capture system\footnote{\url{http://www.vicon.com}} provides reliable state information and a multi-sensor fusion framework~\cite{lynen13robust} fuses the measurement from the onboard IMU in order to compensate for the time delay and low update frequency of the Vicon system. The ascending Technology framework \cite{achtelik2011onboard} is exploited to interface with the vehicle as shown in Fig.~\ref{fig:diagram}. 
\begin{table}[h]
\caption{Quadrotor physical parameters}
\begin{center}
\begin{tabular}{ |c|c| } 
\hline
Parameter & value \\ 
  \hline \hline
mass & \unit[0.665]{kg}\\
dimension & \unit[0.44]{m}, \unit[0.44]{m}, \unit[0.12]{m}\\ 
$I_{xx}$, $I_{yy}$, $I_{zz}$ & \unit[0.007, 0.007, 0.012]{kgm$^2$} \\ 
\hline
\end{tabular}
\end{center}\label{tab:pp}
\end{table}

\begin{figure}
\centering 
\includegraphics[width=0.48\textwidth]{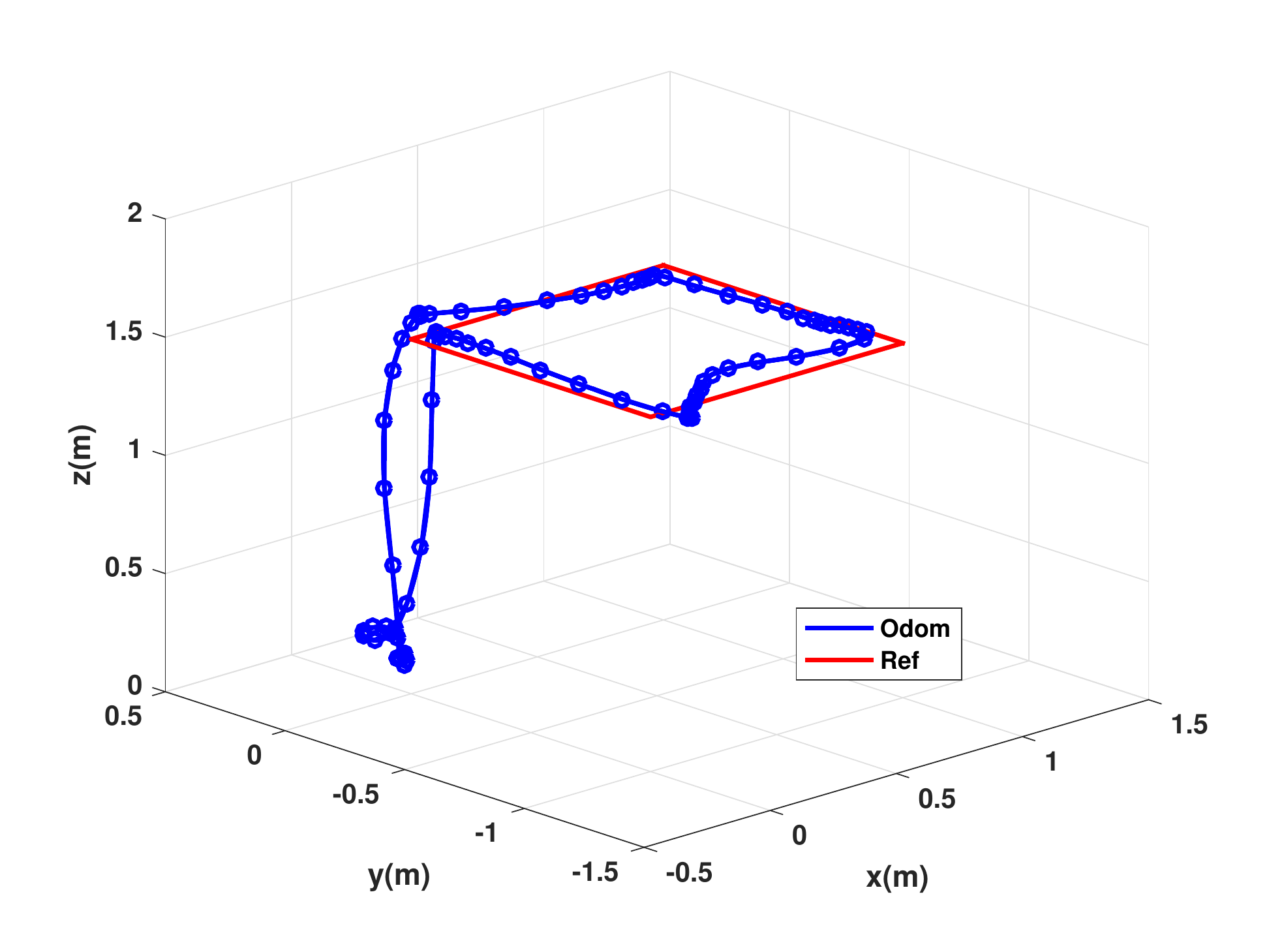}
\caption{Trajectory while performing waypoint tracking.}\label{fig:stepResponse}
\end{figure}

\begin{figure*}
\begin{center}
\begin{minipage}{0.5\textwidth}
\subfigure[Throw 1]{\includegraphics[width=1.0\columnwidth]{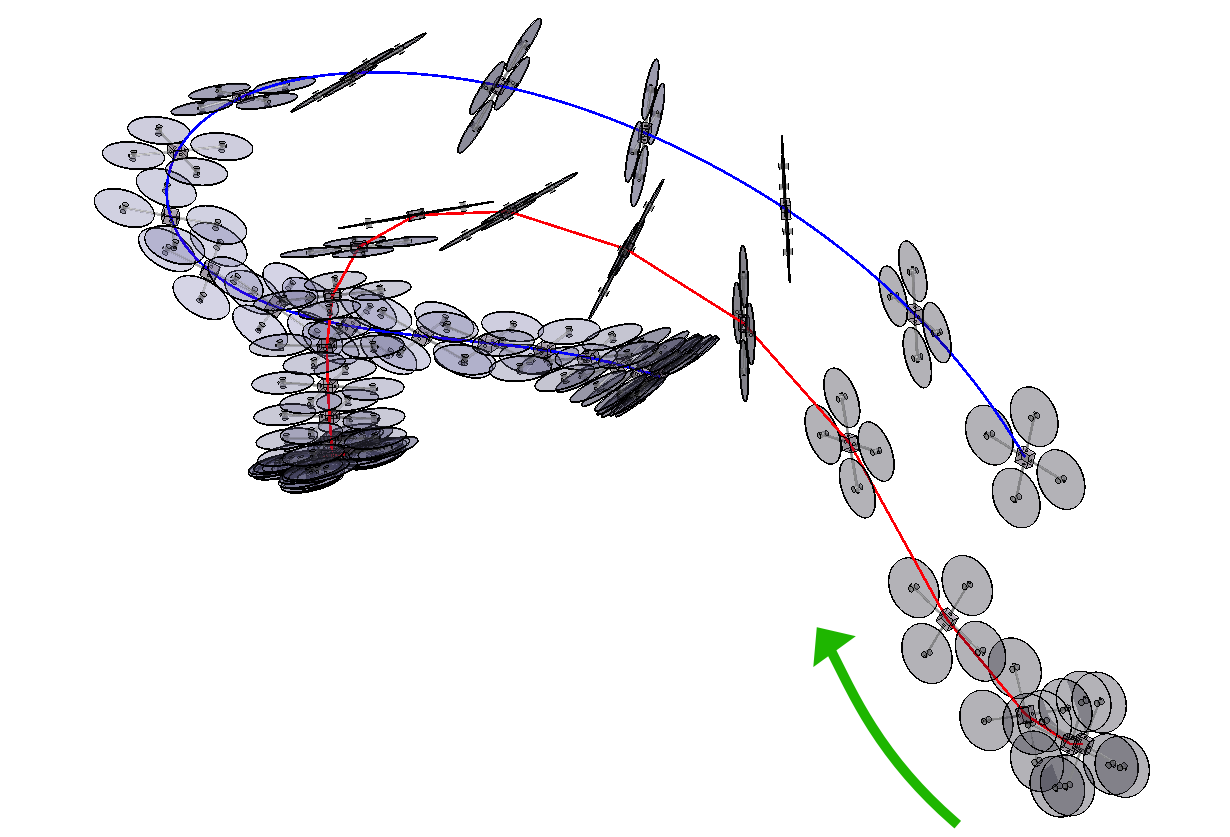}}\label{subfig:throw1}
\subfigure[Throw 2]{\includegraphics[width=1.0\columnwidth]{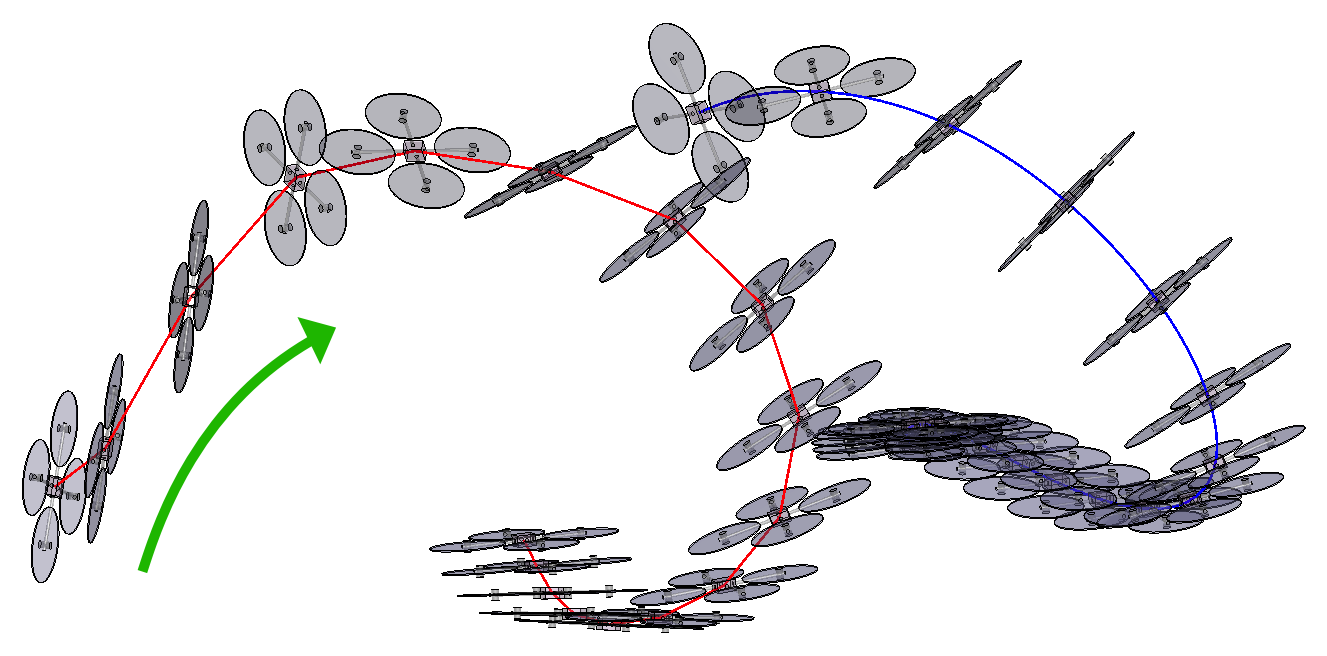}}
\end{minipage}%
\begin{minipage}{0.4\textwidth}
\subfigure[Throw 3]{\includegraphics[width=1\columnwidth]{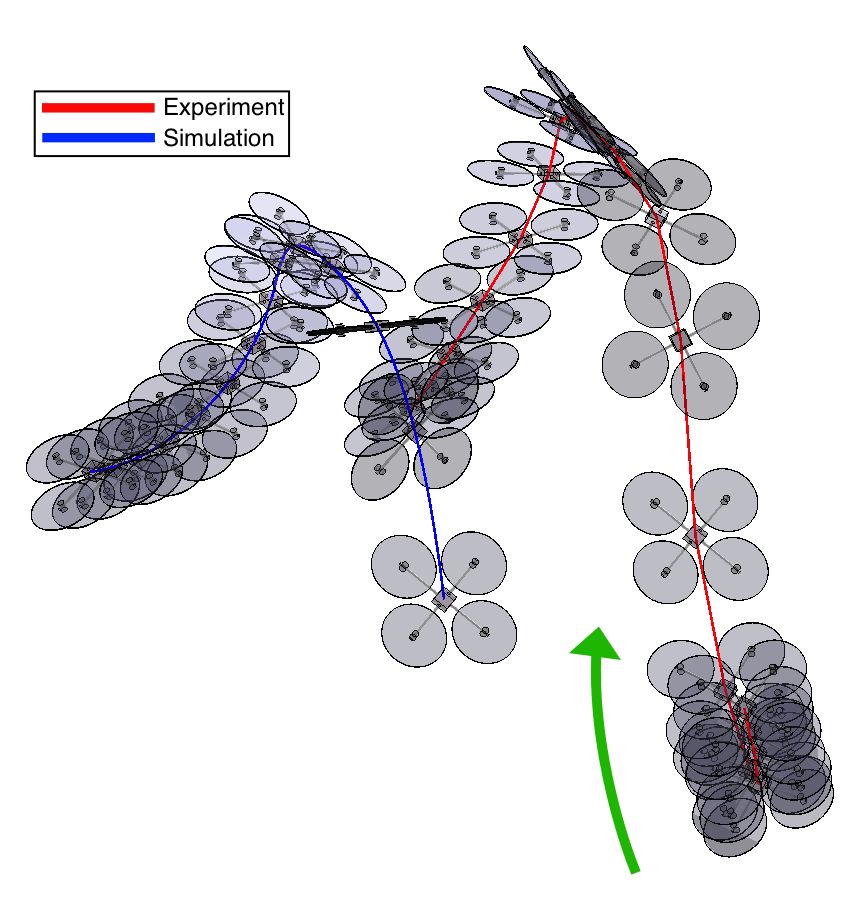}}
\end{minipage}%
\caption{Experimental/simulation results are shown. All trajectories start close to upside-down configuration and the quadrotor flips over in midair. All launches started with initial velocity around \unit[5]{m/s}. The stroboscopic images are generated with \unit[0.1]{s} intervals.}
\label{fig:throws}
\end{center}
\vspace{-0.4cm}
\end{figure*}

We noticed major dynamical differences between simulation and the real quadrotor. First, how the motor controllers regulate the motor speed is unknown. The rotors can be accelerated relatively fast but decelerates quite slowly. We do not get any feedback of the motor speed and the identification of its dynamics is missing at this point. Second, the aerodynamics change significantly near the ground floor. This effect is not modeled to simplify the simulation. Third, we noticed that there are changing dynamic parameters such as battery level and weight distribution~(mostly during battery change). The hovering rotor speed kept changing due to these factors. Fourth, the communication delay in wireless communication and the state estimation error influenced the performance.

We performed a waypoint tracking test with 4 points at the vertices of a \unit[1]{m}-by-\unit[1]{m} square and the result is shown in Fig.~\ref{fig:stepResponse}. There is a minor tracking error but it is not a significant amount. Since the policy was not trained with various external disturbances, it was expected that it has higher tracking error than classical controllers with high gains.

Another test we performed was a manual launch to mimic a recovery scenario when the quadrotor becomes unstable. We manually threw the quadrotor in a very challenging configuration (e.g. upside-down with high linear/angular velocities) and activated the controller when it started falling. In addition, after recording the launch state, we simulated a quadrotor with the same state and the controller. The trajectories from both the experiments and the simulation are shown in Fig.~\ref{fig:throws}. Unintuitively, the quadrotor in the real world showed more stable behavior. We believe that this is due to the unmodeled air drag forces and gyroscopic effect that stabilized the motion of the quadrotor.

One of the trajectories at \unit[45]{deg} is shown in Fig.~\ref{fig:frontImg}. It shows that our policy generates smooth and natural motion while stabilizing the quadrotor at high velocity.

The video clips of all experiments can be found at \url{https://youtu.be/zIi4yHYJdJY}.

\section{Conclusion}\label{sec:conclusion}
We presented a neural-network policy for a quadrotor that is trained in a model-free fashion. Although the simulation is based on the model, not making any use of the model during training frees us from engineering a sophisticated control structure that exploits the model. The trained policy shows outstanding performance and remains computationally cheap at the same time. We also presented a new learning algorithm which outperformed two famous algorithms for this task in terms of computation time. The presented algorithm uses many simulation steps but it is computationally efficient since it minimizes the neural network training steps. It is also a conservative algorithm. There was no issue of divergence during our training which is the main reason why we decided to use it for this project. 

In simulation, we performed way point tracking and recovery tests. We had a small steady state error~(\unit[1.3]{cm}) which can be easily diminished with a constant state offset. It managed to perform waypoint tracking task for extensive time without failure. The tracking error was higher than optimization based controllers. We believe that this is due to the fact that the quadrotor was trained without any disturbances which were present in the real environment. The manual throw test was more successful. The quadrotor was very stable; in fact, it was more stable than what we observed in the simulation. We believe that this is due to the fact that the air drag and gyroscopic effect, which were not present in the simulation, helped to stabilize the system.

Future work will consider ways of introducing more accurate model of the system into the simulation using our parameter estimation techniques~\cite{Sa:2012ICRA}. However, the long term goal is to train RNN which can adapt to errors in modeling automatically. In addition, transfer learning on the real system can further improve the performance of the policy by capturing totally unknown dynamic aspects of the system.




\section*{ACKNOWLEDGMENT}
We thank Mina Kamel (ASL, ETHZ) for providing a software package that can send individual motor rate commands for the experiments.
\bibliographystyle{IEEEtran}
\bibliography{reference}

\end{document}